\documentclass[12pt,a4paper]{article}
\usepackage[utf8x]{inputenc}
\usepackage[T1]{fontenc}
\usepackage{mathptmx} 
\usepackage{multirow}
\usepackage{amsmath}
\usepackage{amssymb}
\usepackage{amsfonts}
\usepackage{setspace}
\usepackage{algorithm2e}

\usepackage{calc} 
\usepackage{enumitem} 

\RequirePackage[authoryear]{natbib}
\RequirePackage[colorlinks,citecolor=blue,urlcolor=blue]{hyperref}
\RequirePackage{graphicx}
\usepackage{bbm, dsfont}
\usepackage{soul}
\usepackage[mathscr]{euscript}
\usepackage{mathrsfs}
\usepackage{booktabs}
\usepackage{url} 
\usepackage{setspace}
\usepackage{multicol}
\usepackage{multirow}
\usepackage{graphicx}
\usepackage{ulem}
\usepackage{amsthm}
\usepackage{rotating} 
\usepackage{mathrsfs}
\usepackage{bbold}

\usepackage{thmtools} 
\usepackage{thm-restate}
\usepackage{booktabs}

\usepackage[a4paper, margin=2cm]{geometry} 

\usepackage[all]{nowidow} 
\usepackage[protrusion=true,expansion=true]{microtype} 

\usepackage[english]{babel}

\newtheorem{theorem}{Theorem}[section]

\newtheorem{definition}[theorem]{Definition}

\newcommand{\bs}{\boldsymbol}

\newcommand{\bbeta}{\boldsymbol{\beta}}
\newcommand{\bdelta}{\boldsymbol{\delta}}
\newcommand{\bX}{\boldsymbol{X}}
\newcommand{\bY}{\boldsymbol{Y}}
\newcommand{\bgamma}{\boldsymbol{\gamma}}
\newcommand{\bGamma}{\boldsymbol{\Gamma}}
\newcommand{\rmd}{\text{d}}
\SetKwData{KwDat}{data}
\newcommand{\RV}{\text{RV}}
\DeclareMathOperator*{\argmax}{arg\,max}

\newtheorem{proposition}{Proposition}[section]

\theoremstyle{plain}

\theoremstyle{remark}

\theoremstyle{empty}

\begin{document}
\doublespacing
\begin{center}
    \large
    \textbf{Canonical Tail Dependence for Soft Extremal Clustering \\ of Multichannel Brain Signals} \footnote{
    Preprint}
        
   \vspace{0.2cm}
    Mara Sherlin D.P. Talento$^{\text{a}*}$,
    Jordan Richards$^{\text{b}**}$, \\
    Rapha\"el Huser$^{\text{c}*}$,
    Hernando Ombao$^{\text{d}*}$
    
    \vspace{0.01cm}
   
    {
    \singlespacing
    \textit{$^a$marasherlin.talento@kaust.edu.sa, $^b$jordan.richards@ed.ac.uk, \\ $^c$raphael.huser@kaust.edu.sa, $^d$hernando.ombao@kaust.edu.sa} \\
    }

    {
    \singlespacing
    $^*$Statistics Program, Computer Electrical and Mathematical Science \& Engineering, King Abdullah University of Science and Technology \\
    $^{**}$ School of Mathematics and Maxwell Institute for Mathematical Sciences, \\ University of Edinburgh, Edinburgh, UK, EH9 3FD \\
    }
    
   \vspace{0.1cm}
    
    \textbf{Abstract}
\end{center}

We develop a novel characterization of extremal dependence between two cortical regions of the brain when its signals display extremely large amplitudes. We show that connectivity in the tails of the distribution reveals unique features of extreme events (e.g., seizures) that can help to identify their occurrence. Numerous studies have established that connectivity-based features are effective for discriminating brain states. Here, we demonstrate the advantage of the proposed approach: that tail connectivity provides additional discriminatory power, enabling more accurate identification of extreme-related events and improved seizure risk management. Common approaches in tail dependence modeling use pairwise summary measures or parametric models. However, these approaches do not identify channels that drive the maximal tail dependence between two groups of signals---an information that is useful when analyzing electroencephalography of epileptic patients where specific channels are responsible for seizure occurrences. A familiar approach in traditional signal processing is canonical correlation, which we extend to the tails to develop a visualization of extremal channel-contributions. Through the tail pairwise dependence matrix (TPDM), we develop a computationally-efficient estimator for our canonical tail dependence measure. Our method is then used for accurate frequency-based soft clustering of neonates, distinguishing those with seizures from those without.

\vfill

\newpage

\section{Introduction} \label{chap:introduction}

Early detection of seizures is crucial, especially at early ages where seizures pose a high risk of disrupting brain development. When unprovoked seizures occur in neonates, they often lead to epileptic encephalopathy \citep{Mizrahi2000neonatal},
hence explaining why close monitoring in neonatal intensive care units (NICUs) is necessary. The primary modality for seizure management in an NICU is the electroencephalogram (EEG) \citep{sandoval2021current}. However, even with medical-grade EEG technology, distinguishing neonates with seizures from those without (i.e., with normal ultrasound recordings) remains challenging, as the patterns and symptoms of seizures can appear deceptively similar \citep{stevenson2019dataset}. While our study is not directly aimed at detecting seizures, we develop a data-driven framework that extracts \textit{unique features} from the tails of the signal distribution which has the ability to isolate, even without expert knowledge, neonates with seizures. Our contributions thus pave the way for efficient representation learning of extremes, which can be useful in a wide range of scientific applications.

Many brain-state analyses involve extraction of features through dependence matrices \citep{tononi1998functional, goutte2001feature, liu2012correlation}. In fact, a number of studies demonstrated associations between physiological signals (through brain functional connectivity), and behavior and cognition \citep{han2013cluster, donofry2020review, Shapson2024Petavoxel}.
In addition to the spectral density matrix \citep[see Ch. 4 of][]{shumway2000time}, canonical coherence---introduced by \cite{brillinger1969canonical} as the time-series analogue of canonical correlation \citep{HHotelling1936}---provides a multivariate measure of synchrony between two groups of signals at specific frequencies. Extending this framework to frequency bands (defined in Section~\ref{subsec:features}) in the presence of outlying observations, the Kendall's tau based canonical coherence \citep{talento2024kencoh} provides a rank-based estimator for the canonical coherence of filtered signals, enabling discrimination of brain states that is robust to outliers.
However, there are cases when connectivity in the bulk of the data differs from connectivity at the tails. As demonstrated in \cite{talento2025spectral}, there are key features of extreme events that are only highlighted when focus is placed on the joint tails of brain signals. 
This is true for brain connectivity, where many extreme neurological events, such as seizures, are a result of concurrent or interrelated extreme firing of neurons \citep{NIH23}.
This complexity has led to recent growing interest in multivariate risk analysis in neuroscience applications \citep{guerrero2023conex, Redondo2024StatExt, talento2025spectral, mcgonigle2025moped}, which addresses questions such as ``What is the likelihood that two brain signals exhibit extreme amplitudes simultaneously?"

The extremal dependence measure (EDM), introduced in \cite{resnick2004extremal} and studied by \cite{ larsson2012extremal}, encodes tail dependence via the framework of multivariate regular variation \citep{resnick2007heavy}. 
\cite{cooley2019decompositions} extended the EDM beyond the bivariate setting by introducing the tail pairwise dependence matrix (TPDM), which, analogous to a covariance matrix, encodes extremal dependence for random vectors. The TPDM has established the basis for a number of multivariate analyses in extreme value theory, such as high-dimensional graphical or causal analysis  \citep{jiang2020principal, gong2024partial, jiang2025separation} and clustering \citep{richards2023modern, elsom2025extreme}. 
In this paper, we utilize the TPDM to propose a novel measure of canonical tail dependence (CTD) to discriminate the joint-tail regional structure of neonates with extreme brain conditions (e.g., seizures) from those patients with normal brain conditions (e.g., seizure-free). We define CTD as the maximized EDM value between linear combinations of signals from two brain regions. Using these optimized linear combinations, we can further investigate patterns associated with different brain conditions based on the tail connectivity of the signals.

A number of studies consider features extracted from multivariate extremes for hard clustering based on, e.g., max-linear models, hidden regular variation, max-stable processes, and conditional limit laws \citep{janssen2020k, meyer2024multivariate, guerrero2022club, o2025clustering}. 
However, extending these hard clustering approaches to multi-subject EEG data is non-trivial. EEG signals display complex and non-stationary cross-dependencies \citep{Redondo2024StatExt, ombao2024spectral}, high noise levels, and significant inter-subject variability \citep{ma2025fcpca}. Thus, relying on a single `hard' cluster in combination with standard feature-extraction techniques frequently fails to represent a subject’s behavior accurately \citep{seghier2007detecting}.
This leads us to develop a novel mechanism for feature extraction from the joint-tails of brain signals and propose an accurate soft clustering framework for multivariate extremes. Our method provides a membership degree to each subject based on their extremal brain connectivity structure. 

One well-known soft clustering technique is fuzzy clustering, grounded in Zadeh’s fuzzy set theory \citep{zadeh1996fuzzy}, which treats partitions between units as continuous transitions rather than hard boundaries. The most widely-used approach is the Fuzzy C-Means (FCM) algorithm \citep{bezdek1984fcm} which has gained popularity in applications that include image \citep{wang2010fast, huang2019brain} and signal processing \citep{murugappan2007eeg, zheng2013fcm, zhang2021anti}. 
There are recent developments in soft clustering of multivariate signals that are quantile-based \citep{lopez2022quantile}, and \cite{ma2025fuzzcoh} showed that canonical coherence-based fuzzy clustering is superior to the state-of-the-art fuzzy clustering, namely, traditional FCM, variable-based principal component analysis (PCA) clustering \citep{he2018unsupervised}, quantile cross-spectral density based fuzzy clustering \citep{lopez2022quantile},  wavelet-based clustering \citep{d2012wavelets}, and fuzzy clustering using PCA \citep{ma2025fcpca}. 
Hence, we apply canonical correlation based soft clustering on seizure data of neonates and presented the result in Section~\ref{chap:exploratory}. From our preliminary analyses, traditional canonical correlation based soft clustering do not adequately separate seizure vs non-seizure patients. 
This is because these clustering methods focus on the bulk of the distribution and are not tailored to extreme events. 
While \cite{d2017fuzzy} proposed a fuzzy clustering of extremes, this is only for univariate time series, and not multivariate brain signals. 

In summary, our paper provides two key innovations: (1) a novel tail dependence measure (the CTD) between random vectors, which facilitates representation learning of extremal connectivity structure; and (2) an unsupervised soft clustering method for multivariate extremes, applied to identification of neonates with seizures. 
Our goal is to provide explainable features that reveal key information contained in the tail of the joint distribution.
Thus, one of the major advantages of our method is its interpretability; our method offers a pathway to relate soft clusters and tail dependence of brain signals. Through the CTD, we can visualize the unifying characteristics of each cluster, facilitating accurate differentiation between neonates with and without seizures. Although applied to brain signals, our method is applicable in other fields, such as finance and environmental science. 

The remainder of the article is organized as follows. Section \ref{chap:exploratory} introduces the neonatal seizure data and an exploratory soft clustering based on classical canonical correlation. 
Section \ref{chap:methodology} details background on multivariate regular variation, presents our novel canonical tail dependence (CTD) measure, and provides our algorithm for fuzzy clustering.
Section \ref{chap:simulation} provides results from a simulation study.
Section~\ref{chap:analysis} reports the novel findings of our application clustering patients in a neonatal intensive care unit through their electroencephalogram recordings. 
Finally, Section~\ref{chap:conclusion} summarizes the study and discusses limitations of the proposed method.


\section{The Neonatal Electroencephalography Data} \label{chap:exploratory}

In this section, we first introduce the dataset (Section \ref{subsec:data}) and then describe the EEG preprocessing steps used to address the spectral-domain analysis problem (Section \ref{subsec:features}). In Section \ref{subsec:CCA-clustering}, we present a brief exploratory analysis using canonical-correlation-based feature extraction.

\subsection{Data} \label{subsec:data}

The data are available at \url{https://zenodo.org/records/2547147} and the EEG signals were recorded from neonates admitted to an intensive care unit in Finland \citep{stevenson2019dataset}. These signals were recorded at $19$ electrodes covering the entire scalp.
Seizure occurrence labels are based on expert annotations.
We follow the preprocessing pipeline of \cite{talento2025spectral}: selecting electrodes with consistent quality across subjects, followed by application of the Early Stage Pre-processing (PREP) pipeline \citep[see][for details]{bigdely2015prep}. PREP is chosen over alternative artifact detection methods as it preserves extremal behavior. 
Subjects with an excessive proportion of suspiciously small values in the cleaned EEG are excluded.
The analysis includes $N = 14$ subjects: six non-epileptic subjects and eight epileptic subjects with seizures primarily localized in the right hemisphere of their brain.


\begin{figure}
        \centering
        \includegraphics[width=1\linewidth]{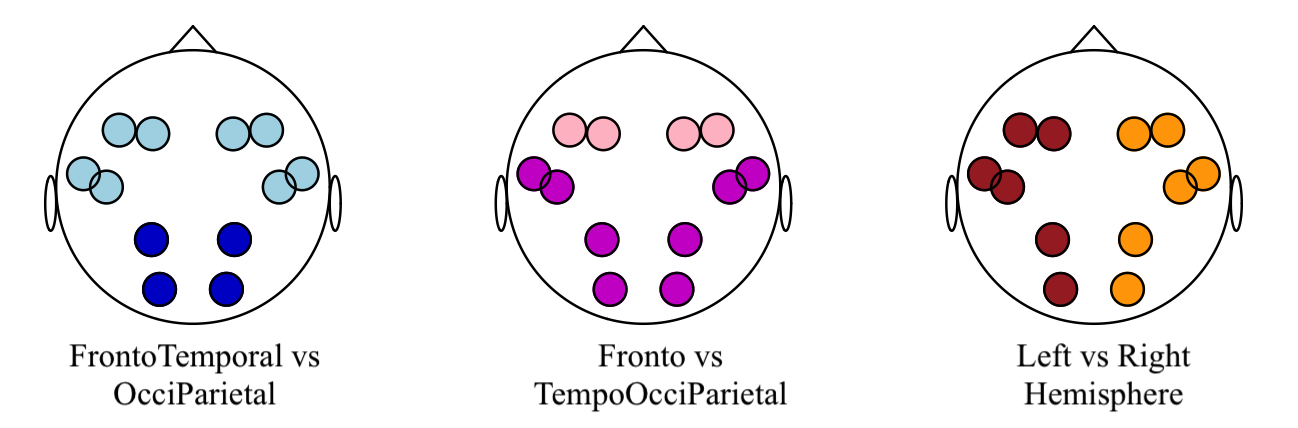}
        \caption{EEG channel locations on the scalp of different regions of interests: FrontoTemporal (light blue nodes) vs OcciParietal (dark blue nodes), Frontal (light pink nodes) vs TempoOcciParietal (dark pink nodes), and left (brown nodes) vs right hemisphere (orange nodes).}
        \label{Fig:EEG-nodes}
\end{figure}

\begin{figure}
        \centering
        \includegraphics[width=1\linewidth]{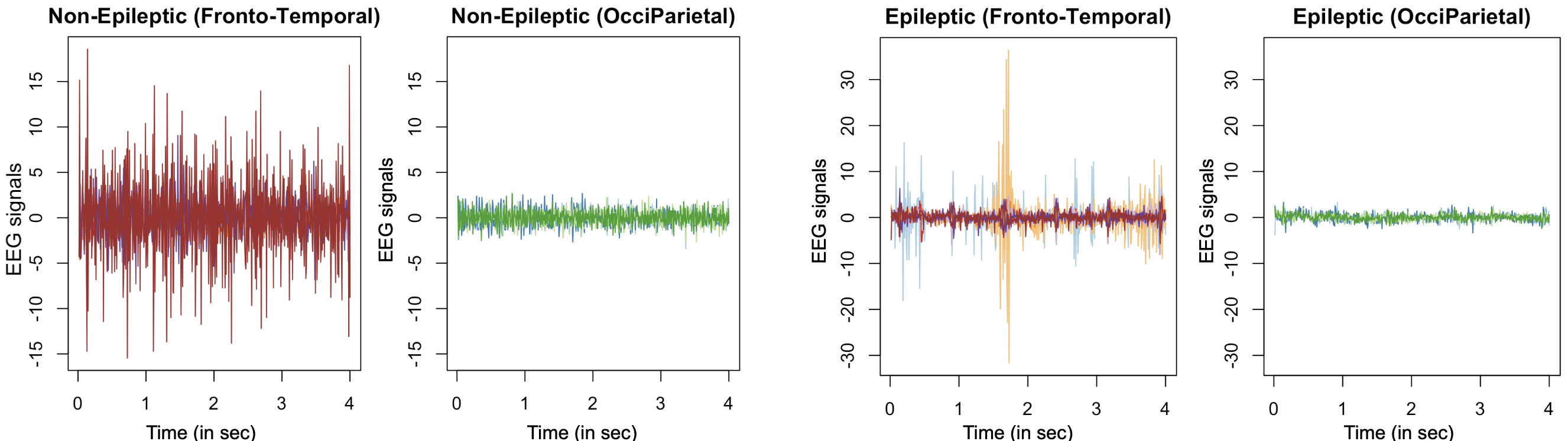}
        \caption{Detrended EEG signals from FrontoTemporal lobe (1st and 3rd panels) and OcciParietal lobe (2nd and 4th panels) of non-epileptic patient (1st and 2nd panels) and epileptic patient (3rd and 4th panels).}
        \label{Fig:EEG-signals}
\end{figure} 

We seek to discern seizure-free neonates vs neonates who had primary localization of seizure at the right hemisphere. 
To do this, two groups of channels associated with seizure activity were identified, denoted as $\bX \in \mathbb{R}^P$ and $\bY \in \mathbb{R}^Q$ (defined more formally in Equation~\eqref{eq:CTD}).
Three forms of region-to-region connectivity based on $D = P+Q = 12$ channels are of interest in our analysis, namely:

\noindent \textbf{(i) FrontoTemporal-vs-OcciParietal (FT-OP): } \ (F3, F7, F4, F8, T3, T4, T5, T6) vs (P3, P4, O1, O2) i.e., $P = 8$ and $Q = 4$. Most primary localizations in the subjects arise from the OcciParietal lobe; thus, isolating these channels provides a more informative spatial configuration for our analysis. Moreover, literature highlights the temporal lobe as the most common focal point of seizures \citep{Pisani2021}. See blue colored nodes in Figure~\ref{Fig:EEG-nodes} (left panel). 

\noindent \textbf{(ii) Fronto-vs-TempoOcciParietal (F-TOP): } \ (F3, F7, F4, F8) vs (T3, T4, T5, T6, P3, P4, O1, O2) i.e., $P = 4$ and $Q = 8$. The frontal lobe is known to play a role in motor behavior in neonates, which may be disrupted by seizures \citep{panayiotopoulos2005neonatal}. Hence, we tested the partitioning frontal lobe from all other lobes. See pink colored nodes in Figure~\ref{Fig:EEG-nodes} (middle panel). 

\noindent \textbf{(iii) Left-vs-Right Hemisphere: } \  (F3, F7, T3, T5, P3, O1) vs (F4, F8, T4, T6, P4, O2) i.e., $P = 6$ and $Q = 6$. As the primary localization of the seizures are in the right hemisphere, we also explore the connectivity between the left and right hemispheres. See orange and brown colored nodes in Figure~\ref{Fig:EEG-nodes} (right panel). 

Figure~\ref{Fig:EEG-signals} shows sample signals taken from a non-epileptic and epileptic patients. From here, we can see that FrontoTemporal channels (1st and 3rd panels) are heavy-tailed for both epileptic and non-epileptic patients. However, the tail behavior seems different. The goal of our paper, then, is to provide a practical feature extraction tool for examining potential discrepancies in tail behavior across brain conditions.

\subsection{Spectral Domain Features} \label{subsec:features}

Analyzing brain signals during high-frequency oscillations offers information on bursts of its electrical activity, e.g., seizures \citep{guerrero2023conex, pinto2024connectivity}. 
We thus extract features from the EEGs that are attributed to the standard frequency bands: delta, for $\Omega = (0 - 4]$Hz; theta, for $\Omega = (4 - 8]$Hz; alpha, for $\Omega = (8 - 12]$Hz; beta, for $\Omega = (12 - 30]$Hz; gamma, for $\Omega = (30 - 50]$Hz. 

Our approach is to divide the EEG recording into $B$ disjoint blocks where each block contains $A$ number of time points. 
For block $b$, denote the local discrete Fourier transform (DFT) of the $j$-th channel at the fundamental frequency $\omega_a := a/A$, $a = 0,1, \dots, A-1$, as
\begin{equation*}
    d_{j,b}(\omega_a) = \frac{1}{\sqrt{A}} \sum_{t=A(b-1)+1}^{A(b-1)+A} E_{j,t} \exp(-\text{i}2\pi \omega_at), \ \ \text{for } j = 1, \dots, D,
\end{equation*}
where $-\text{i} = \sqrt{-1}$ and $\bs{E}_t \in \mathbb{R}^D$ denotes the EEG signals observed at time $t$, $t=1,2,\dots,AB$, that is `locally stationary': the time series is approximately stationary on each of the $B$ time blocks.
Then, we compute the local periodogram (at block $b$ and frequency $\omega_a$),
\begin{equation*}
    Z_{j,b}(\omega_a) = |d_{j,b}(\omega_a)|^2 = \frac{1}{A} \bigg| \sum_{t= A(b-1)+1}^{A(b-1)+A} E_{j,t} \exp(-\text{i}2\pi \omega _at) \bigg| ^2.
\end{equation*}
The local periodogram $Z_{j,b}(\omega_a)$ measures the contribution of the amplitude of the oscillation at frequency $\omega_a$ to the total variance.
Consistent with standard analyses of brain signals \citep{Srinivasan2007EEG}, we aggregate the local periodogram over the standard frequency bands. Hence, the local $\Omega$-periodogram for channel $j = 1, \dots D$, at time-block $b$, is
\begin{equation} \label{eq:LocalBandPeriodogram}
    Z_{j,b}(\Omega) = \frac{1}{|\Omega|} \sum_{\{a: \mbox{SR}\times\omega_a \in \Omega \}} Z_{j,b}(\omega_a),
\end{equation}
where SR is the sampling rate.
In our application, the sampling rate is SR = 256Hz and the local time window is two seconds; hence, $A = 256 \times 2 =  512$.

\subsection{Canonical correlation based soft clustering} \label{subsec:CCA-clustering}

\begin{figure}
        \centering
        \includegraphics[width=1\linewidth]{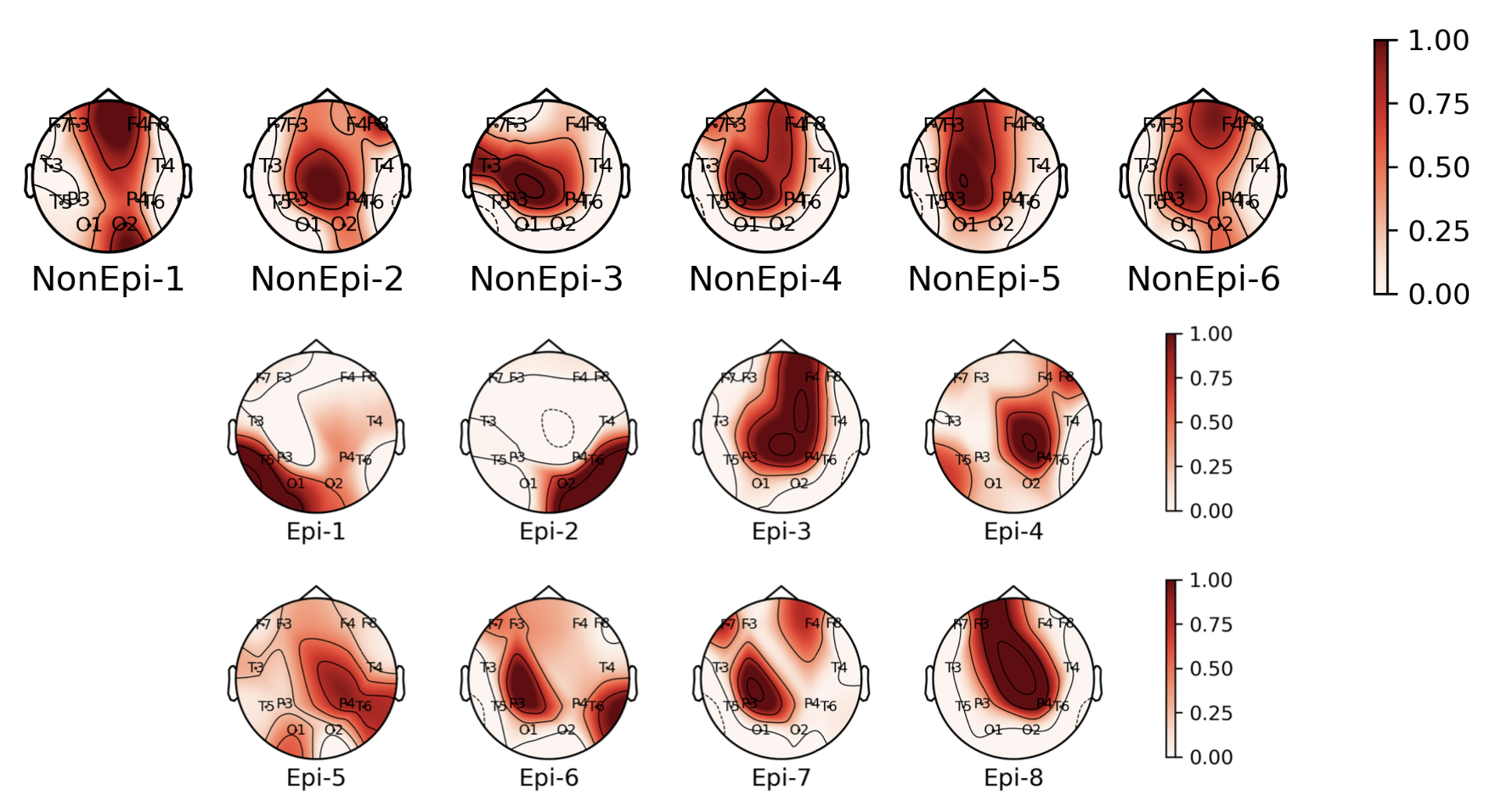}
        \caption{Tomomaps \citep[using \texttt{mne} Python package, see][]{Gursoy2014Tomopy} of $|\hat{\bs{\Lambda}}^{(n)}_0|$, at the gamma band for FrontoTemporal-vs-OcciParietal connectivity. Canonical vectors of six non-epileptic neonates (first row) and eight epileptic neonates (second and third row) with primary localization on \textit{right hemisphere}. Darker color represents higher eigenvector magnitude for channels $j = 1, \dots, 12$, labeled F3, F7, F4, F8, T3, T4, T5, T6, P3, P4, O1, O2.}
        \label{Fig:CCA-topo}
\end{figure}

Canonical correlation analysis (CCA) finds the maximum linear association among all possible linear combinations of the columns of two matrices. Let $\bs{X} \in \mathbb{R}^P$ and $\bs{Y} \in \mathbb{R}^Q$. The
CCA finds the vectors $\bs{\gamma}_0 \in \mathbb{R}^P$ and $\bs{\beta}_0 \in \mathbb{R}^Q$ that maximize the correlation between $\bs{\gamma}_0^\top \bX$ and $\bs{\beta}_0^\top \bY$. These vectors are called ``canonical directions" or ``canonical variates"; and the corresponding correlation is the ``canonical correlation". Precisely, the canonical correlation is defined by \cite{HHotelling1936} as
\[\rho = \max_{\bs{\gamma}_0, \bs{\beta}_0} \text{Cor}(\bs{\gamma}_0^\top \bX, \bs{\beta}_0^\top \bY)^2 \in (0,1),\]
subject to the constraint $\bs{\gamma}_0^\top \text{Cov}(\bX, \bX) \bs{\gamma}_0 = \bs{\beta}_0^\top \text{Cov}(\bY, \bY) \bs{\beta}_0 = 1$ for identifiability.

\begin{figure}
        \centering
        \includegraphics[width=1\linewidth]{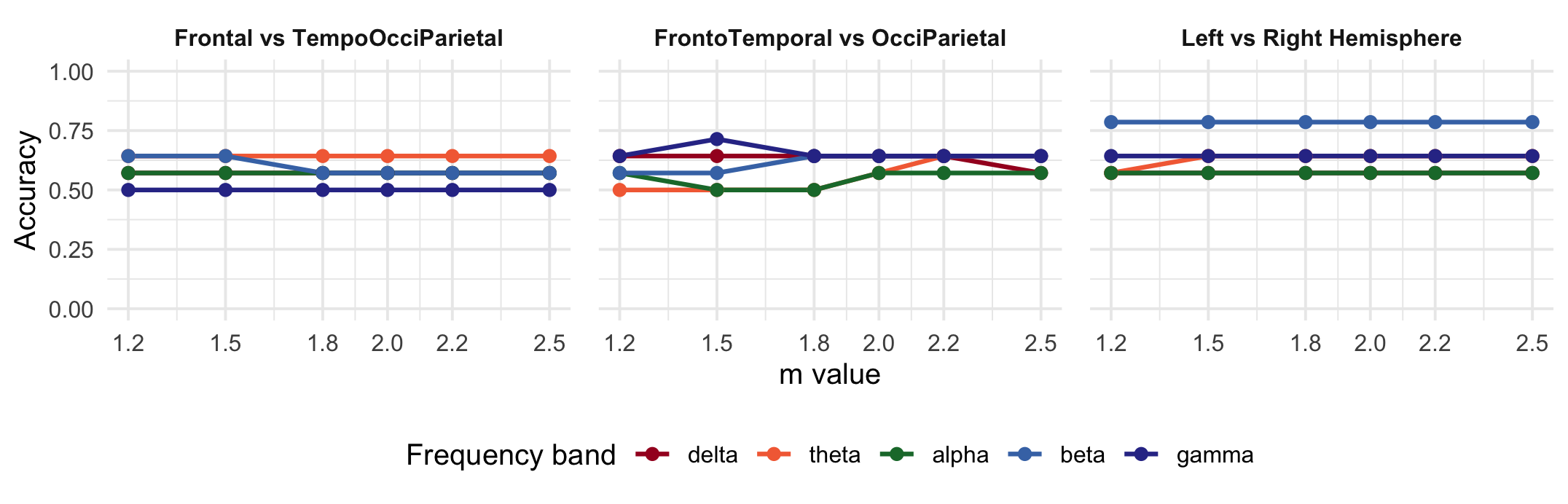}
        \caption{Accuracy of CCA-based fuzzy clustering evaluated for different values of fuzzy parameter $m$ and for all frequency bands (i.e., delta, theta, alpha, beta and gamma). Colors differ with frequency band and the regions of interest differ across panels.}
        \label{Fig:CCA-RI}
\end{figure}

Define $\bs{Z}_b^{(n)} = (\bX^{(n)\top}_b,\bY^{(n)\top}_b)^\top$ as the concatenated vector of $\bs{X}_b \in \mathbb{R}^P$ and $\bs{Y}_b \in \mathbb{R}^Q$, for $b = 1, \dots, B$ blocks and $n = 1, \dots, N$ subjects. For notational simplicity, we omit $\Omega$ in Equation~\eqref{eq:LocalBandPeriodogram}, as the analyses are carried out separately for each frequency band.
For $N = 14$, we obtain estimates of ${\bs{\Lambda}_0^{(n)} := (\bs{\gamma}_0^{(n)\top}, \bs{\beta}_0^{(n)\top})^\top \in \mathbb{R}^{P+Q}}$, through the estimation procedure provided in Chapter 10 of \cite{Mardia1979bibby}. We apply the fuzzy C-means algorithm to absolute values of $\bs{\Lambda}_0 := (\bs{\Lambda}_0^{(1)}, \dots, \bs{\Lambda}_0^{(N)})^\top \in \mathbb{R}_+^{N \times D}$, where $D = P+Q$, with two clusters (to identify epileptic and non-epileptic subjects).
Here, we use $m$ to indicate the degree of fuzziness, with higher values of $m$ producing softer clustering assignments. A more formal definition of $m$ is to be provided in Section~\ref{chap:subsec:fuzzyclust}. 

We display (smoothed) estimates of $|\bs{\Lambda}_0|$ for FrontoTemporal-vs-OcciParietal connectivity in Figure~\ref{Fig:CCA-topo}, i.e., $|\bs{\Lambda}_0| \in \mathbb{R}_+^{14 \times 12}$. From this figure, it is unclear how the connectivity structure of the spectral features (i.e., periodogram) changes for different clusters. Figure~\ref{Fig:CCA-RI} provides the clustering accuracy (see Equation~\eqref{eq:accuracy} below for its exact definition) of the CCA-based clustering procedure for different levels of $m$, different frequency bands and the three regional connectivities. The limitation of the standard CCA is that the accuracy is not sufficiently high (under 80\%) to distinguish epileptic and non-epileptic subjects. This is a major weakness and is a motivation to develop a feature extracting mechanism tailored to extreme events. In the next section, we develop a novel method for the joint tails of the distribution using multivariate extreme value theory. Our method provides a principled way of extracting connectivity-based features from the tails of the distribution to offer an interpretable and accurate way of clustering brain signals.


\section{The Proposed Methodology} \label{chap:methodology}

We detail the multivariate regular-variation framework in Section~\ref{subsec:background}, and then introduce our novel canonical tail dependence (CTD) measure in Section~\ref{subsec:methods:CTD}.
This is followed by our proposed fuzzy extremal clustering algorithm in Section~\ref{chap:subsec:fuzzyclust}. 

\subsection{Background: Multivariate Regular Variation} \label{subsec:background}

A well-established representation of joint tail behavior is multivariate regular variation \citep{resnick2007heavy}. We follow \cite{Dan2025workingdraft} in its definition.

\begin{definition} \label{def:MRV}
    {\citep[\textit{Multivariate Regular Variation,}][]{Dan2025workingdraft}} \newline 
    A $P$-dimensional random vector, $\bs{X}$, is `regularly varying' of order $\alpha > 0$, i.e., ${\bs{X} \in \text{RV}^P(\alpha)}$, if there exists a normalizing sequence $w(u) \rightarrow \infty$ such that
        \begin{equation*}
            w(u)\mathbb{P}(u^{-1}\bs{X} \in C) \overset{\mathrm{v}}{\rightarrow} \nu_{\bs{X}}(C), \; \; \; u \rightarrow \infty,
        \end{equation*}
where $\nu_{\bs{X}}(\cdot) \in [-\infty,\infty]^P \backslash \{\bs{0}_{P}\} := \overline{\mathbb{R}}^P$ is a limiting measure on the space of (non-trivial) Radon measures for Borel sets $C \in [-\infty,\infty]^P \backslash \{\bs{0}_{P}\},$ and $\overset{\mathrm{v}}{\rightarrow}$ denotes `vague' convergence of measures. The limit measure is $(-\alpha)-$homogeneous, i.e., $\nu_{\bs{X}}(zC) =  z^{-\alpha}\nu_{\bs{X}}(C)$, and the normalizing sequence satisfies $\lim_{u \rightarrow \infty} w(u z)/w(u) = z^{\alpha}$, for $z>0$. 
\end{definition}

Following \cite{cooley2019decompositions}, we set $\alpha = 2$. This is without loss of generality (with respect to marginal transformations) as we can always standardize the data to have $\text{RV}^1(2)$ margins \citep[see, e.g.,][]{jiang2025efficient}.
The homogeneity of the limit measure, $\nu_{\bs{X}}(\cdot)$, makes a radial-angular representation particularly well-suited for describing multivariate regular variation. Let $(R,\bs{\psi}) := ( \|\bs{x}\| , \bs{x} / \|\bs{x}\|)$ be the vector of radial and angular components where $\|\cdot\|$ is the L$_2$ norm. For large $r > 0$, define the set
\begin{equation}
    \label{eq:C_def}
C(r,\Psi) := \{\bs{x} \in \overline{\mathbb{R}}^P : R > r , \bs{\psi} \in \Psi\},
\end{equation}
where $\Psi \subset \mathbb{S}^{P-1}$, with $\mathbb{S}^{P-1} := \{\bs{x} \in \mathbb{R}^P : \|\bs{x}\| = 1\}$ the unit $(P-1)$-sphere. We then define the \textit{angular measure} as
$H_{\bs{X}}(\Psi) := \nu_{\bs{X}}(C(1,\Psi))$, which  allows us to quantify the relative contribution of components of $\bX$ to multivariate extreme events \citep{Dan2025workingdraft}. In other words, $H_{\bs{X}}$ can quantify the strength of extremal association between variables through $\bs{\psi}$ \citep[see][for details]{resnick2007heavy}. 
\cite{resnick2004extremal} provide a summary measure of pairwise extremal association by defining the extremal dependence measure (EDM) of a vector $(X_1, X_2)^\top \in \RV^2(2)$ as
        \[\sigma(X_1, X_2) = \int_{\mathbb{S}^{1}} \psi_1 \psi_2 \text{d}H_{(X_1,X_2)}(\bs{\psi}) = 2\lim_{r \rightarrow \infty} \mathbb{E}(\psi_1\psi_2 | R > r) \in [0,1],\]
where the strength of extremal dependence in $(X_1,X_2)^\top$ increases with $\sigma$.
\cite{cooley2019decompositions} proposed the tail pairwise dependence matrix (TPDM) which extends the EDM to a $P \times P$ matrix. We denote the TPDM by $\bGamma \in \mathbb{R}^{P \times P}$ and its elements are given by the following formula:
\[{\Gamma}_{jk} :=  \int_{\mathbb{S}^{P-1}} \psi_j \psi_k \text{d}H_{\bs{X}}(\bs{\psi}) = P\lim_{r \rightarrow \infty} \mathbb{E}(\psi_j \psi_k | R > r) \]
for $j,k = 1, \dots, P$, where now the radius $R$ and the angles $\psi_j, \psi_k$ are computed based on the full random vector $\bs{X} \in \mathbb{R}^P$. Furthermore, note that (a) the TPDM evaluates the integral over the entire set $\mathbb{S}^{P-1}$, in contrast to the set $\mathbb{S}^{1}$ used in the EDM, and (b) the TPDM is a positive semi-definite and symmetric matrix. 

\subsection{Canonical Tail Dependence}  \label{subsec:methods:CTD}

Consider now a $D$-dimensional regularly varying vector $\bs{Z} \in \RV^D(2)$. Suppose that $\bs{Z}$ is composed of two distinct subsets, so that $\bs{Z} = (\bs{X}^{\top}, \bs{Y}^{\top})^{\top}$, where ${\bs{X} \in \RV^P(2)}$ represents features from one brain region, and 
$\bs{Y}  \in \RV^Q(2)$ represents features from a second brain region, and ${D = (P + Q) }$ with $P,Q\geq 2$. We summarize extremes in each brain region using vectors $\bs{\gamma} \in \mathbb{R}^P$ and $\bbeta \in \mathbb{R}^Q$. Let $\bs{\gamma}^{\top}\bX \in \mathbb{R}$ and $\bbeta^{\top}\bY \in \mathbb{R}$ be the projections of $\bs{X}$ and $\bs{Y}$ onto the line defined by vectors $\bgamma$ and $\bbeta$ for brain regions one and two, respectively. The canonical tail-dependence is the squared maximal EDM between the projections of the two brain regions. 

\begin{definition} \label{def:CTD}
    Let $\bs{X} \in \text{RV}^P(2)$, $\bs{Y} \in \text{RV}^Q(2)$, $\bgamma \in \mathbb{R}^P$, and $\bs{\beta} \in \mathbb{R}^Q$. The ``canonical tail dependence'' (CTD) is 
    \begin{eqnarray}
        \tau &=&  \max_{\bgamma, \bbeta} \left\{ \sigma\left( 
           \bs{\gamma}^{\top}\bX, \ \ \bs{\beta}^{\top}\bY\right) \right\}^2  \in [0,1].\label{eq:CTD}
    \end{eqnarray}
\end{definition}

The canonical tail dependence, $\tau$, 
generalizes the approach in \cite{russell2016data}, which regresses multiple extreme covariates on a single dependent variable. With the CTD, our goal is to identify linear combinations of two sets of variables that maximize tail dependence.
Direct numerical maximization of this objective function, however, incurs a substantial computational cost, particularly for large $P$ and $Q$. 
To drastically simplify computations, we derive an analytical solution to Equation~\eqref{eq:CTD} using the tail pairwise-dependence matrix (TPDM).


\begin{proposition} \label{prop:solnTPDM}
    Let $\bs{Z} = (\bX^\top, \bY^\top)^\top$ where $\bs{X} \in \text{RV}^P(2)$ and $\bs{Y} \in \text{RV}^Q(2)$. Define $\bs{\Gamma}_{XX} \in \mathbb{R}^{P \times P}$ and $\bs{\Gamma}_{YY} \in \mathbb{R}^{Q \times Q}$ to be the TPDMs of $\bX$ and $\bY$, respectively. Moreover, let $ \bs{\Gamma}_{XY} = \bs{\Gamma}_{YX}^\top \in \mathbb{R}^{P \times Q}$ be the cross-TPDM between $\bX$ and $\bY$.
    The TPDM of $\bs{Z}$ has the block matrix form
    \[\bs{\Gamma} = \begin{pmatrix}
        \bs{\Gamma}_{XX} & \bs{\Gamma}_{XY} \\
        \bs{\Gamma}_{YX}  & \bs{\Gamma}_{YY} 
    \end{pmatrix}.\]
    The CTD in Equation~\eqref{eq:CTD} is equivalent to
    \begin{gather*}
        \tau  = \max_{ 
                \substack{\bgamma, \bbeta: \\
                 \bs{\gamma}^{\top} \bs{\Gamma}_{XX}\bs{\gamma} = \bs{\beta}^{\top} \bs{\Gamma}_{YY}\bs{\beta} = 1}}  
                 \{ \bs{\gamma}^{\top}  \bs{\Gamma}_{XY}\bs{\beta} \}^2,
    \end{gather*}
    
\end{proposition}

\begin{proof}
    The proof of this proposition takes advantage of the linearity of the EDM \citep[Proposition 1,][]{Dan2025workingdraft}.
We show that ${\sigma\left( \bs{\gamma}^{\top}\bX, \ \ \bs{\beta}^{\top}\bY \right)=\bs{\gamma}^{\top}  \bs{\Gamma}_{XY}\bs{\beta}}$. 
Define ${\zeta : \overline{\mathbb{R}}^{P+Q} \rightarrow \overline{\mathbb{R}}^2}$ as the function 
$$\zeta(\bs{Z}) = (\bs{\gamma}^{\top}\bX , \bs{\beta}^{\top}\bY) := \bs{Z}_\zeta.$$
We then define the \textit{canonical angles} $\bs{\psi}_\zeta$ (similarly to Equation~\eqref{eq:C_def}) as 
\[ \bs{\psi}_\zeta:=(\psi_{\zeta,1},\psi_{\zeta,2})^\top =\bs{Z}_\zeta / \|\bs{Z}_\zeta\| \in \Psi_\zeta \subset \mathbb{S}^1, \]
which can be linked backed to the angles $\bs{\psi}$ of $\bs{Z}$ through the transformation ${\bs{\psi}_{\zeta} :=\zeta(\bs{\psi})} / \|\zeta(\bs{\psi})\|$. 
Further, let $\zeta^{-1}(\Psi_\zeta) := \{ \bs{\psi} \in\mathbb{S}^{D-1} : \zeta(\bs{\psi}) / \|\zeta(\bs{\psi}) \in \Psi_{\zeta}\}$.
Following \cite{Dan2025workingdraft}, the limit measure of $\bs{Z}$ can be rewritten in terms of the conditional set
\begin{equation*}
    C(1,\Psi_\zeta) := \{ \bs{Z} \in \overline{\mathbb{R}}^D :\|\zeta(\bs{\psi})\| > 1, \zeta(\bs{\psi}) /\|\zeta(\bs{\psi})\| \in \bs{\Psi}_\zeta \}, 
\end{equation*}
which depends only on the linear combination, $\bs{Z}_\zeta$. Hence, the angular measure $H_{\bs{Z}_\zeta}$ of $\bs{Z}_\zeta$ is now given as 
\begin{eqnarray}
    \nu_{\bs{Z}}(C(1,\Psi_\zeta)) &=& \int_{\zeta^{-1}(\Psi_\zeta)} \int_{1/\|\zeta(\bs{\psi})\|}^\infty \alpha r^{-\alpha - 1} \text{d}r\text{d}H_{\bs Z}(\bs{\psi}) \notag \\
    &=&  \int_{\zeta^{-1}(\Psi_\zeta)}\|\zeta(\bs{\psi})\|^{\alpha} \rmd H_{\bs{Z}}(\bs{\psi}) = :H_{\bs{Z}_\zeta}(\bs{\psi}_\zeta) \label{eq:Hz}.
\end{eqnarray}


\noindent We partition the set $\{\bs{\psi} \in \mathbb{S}^{D-1} : \zeta(\psi)/\|\zeta(\bs{\psi})\| \in \Psi_\zeta\}$ to correspond with the indices of $\bX$ and $\bY$, i.e., $\bs{\psi} := (\bs{\psi}_X^\top, \bs{\psi}_Y^\top)^\top$. Hence, by this definition and $\alpha = 2$, we have

\begin{eqnarray*}
    \sigma\left(  \bs{\gamma}^{\top}\bX, \ \ \bs{\beta}^{\top}\bY\right) &=& \int_{\mathbb{S}^1} \psi_{1,\zeta}\psi_{2,\zeta} \ \rmd H_{\bs{Z}_\zeta}(\bs{\psi}_\zeta) \\
    &=& \int_{\mathbb{S}^{D-1}} \frac{\bs{\gamma}^\top \bs{\psi}_X}{\|\zeta(\bs{\psi})\|} \frac{\bs{\beta}^\top \bs{\psi}_Y}{\|\zeta(\bs{\psi})\|} {\|\zeta(\bs{\psi})\|}^2 \rmd H_{\bs{Z}}(\bs{\psi}) \\
    &=& \int_{\mathbb{S}^{D-1}}  (\gamma_1\psi_1 + \dots + \gamma_P\psi_P) (\beta_1\psi_{P+1} + \dots +\beta_Q\psi_{D}) \rmd H_{\bs{Z}}(\bs{\psi}) \\
    &=& \gamma_1\int_{\mathbb{S}^{D-1}}  \psi_1\psi_{P+1} \rmd H_{\bs{Z}}(\bs{\psi}) \beta_1 + \dots + \gamma_P\int_{\mathbb{S}^{D-1}}  \psi_P\psi_{D} \rmd H_{\bs{Z}}(\bs{\psi}) \beta_Q \\
    &=& \gamma_1 \Gamma_{1,P+1} \beta_1 + \dots + \gamma_P\Gamma_{P,D} \beta_{Q} \\
    &=& \bs{\gamma}^{\top}  \bs{\Gamma}_{XY}\bs{\beta},
\end{eqnarray*}
where the second line follows from Equation~\eqref{eq:Hz} and the transformation $\bs{\psi}_{\zeta} := \zeta(\bs{\psi}) / \|\zeta(\bs{\psi})\|$. Note that the surface Jacobian of this transformation is equal to one. 
\end{proof}

Proposition~\ref{prop:solnTPDM} gives a straightforward estimator of the CTD. After estimating the TPDM, $\bGamma$, we solve the optimization problem in Equation~\eqref{eq:CTD} through Proposition~\ref{prop:solnTPDM}, i.e., the optimization of CTD reduces to an eigen-decomposition problem similar to Theorem 10.2.1 in \cite{Mardia1979bibby}.
We then employ this optimization problem to characterize a novel connectivity structure in the joint extremes of two vectors, which we call ``canonical tail variates'' (defined below).
\begin{definition} \label{def:CTD-variates}
    Let $\bs{X} \in \text{RV}^P(2)$, $\bs{Y} \in \text{RV}^Q(2)$, $\bgamma \in \mathbb{R}^P$, and $\bs{\beta} \in \mathbb{R}^Q$. Further, let $\bs{\Gamma}_{XX} \in \mathbb{R}^{P \times P}$ and $\bs{\Gamma}_{YY} \in \mathbb{R}^{Q \times Q}$ be the TPDMs of $\bX$ and $\bY$, respectively. The ``canonical tail-variates'', denoted by $\bgamma^*, \bbeta^*$, are defined as
    \begin{equation}
         \label{eq:CT-variates}
        \bgamma^*, \bbeta^*  = \argmax_{\bgamma, \bbeta} \left\{ \sigma\left( 
           \bs{\gamma}^{\top}\bX, \ \ \bs{\beta}^{\top}\bY\right) \right\}^2,\\
    \end{equation}
    subject to the constraint $\bs{\gamma}^{\top} \bs{\Gamma}_{XX}\bs{\gamma} = \bs{\beta}^{\top} \bs{\Gamma}_{YY}\bs{\beta} = 1$ for identifiability.
    Furthermore, the ``extremal scores'' are defined as $\bgamma^{*\top}\bX$ and $\bbeta^{*\top}\bY$.
\end{definition}

\noindent From Definition~\ref{def:CTD-variates} and Proposition~\ref{prop:solnTPDM}, it follows that the canonical tail-variates can be computed as
    \[ \bgamma^*, \bbeta^* = \argmax_{
                        \substack{\bgamma, \bbeta: \\
                        \bs{\gamma}^{\top} \bs{\Gamma}_{XX}\bs{\gamma} = \bs{\beta}^{\top} \bs{\Gamma}_{YY}\bs{\beta} = 1}}  
                        \{ \bs{\gamma}^{\top}  \bs{\Gamma}_{XY}\bs{\beta} \}^2. \]
                        
\noindent Theorem 10.2.1 in \cite{Mardia1979bibby} further implies that ${\tau}$ is the first eigenvalue of
\begin{equation}
    \bs{G}_1 = {\bGamma}_{XX}^{-1}{\bGamma}_{XY}{\bGamma}_{YY}^{-1}{\bGamma}_{YX} \ \ \text{or, equivalently, of} \ \ \bs{G}_2 = {\bGamma}_{YY}^{-1}{\bGamma}_{YX}{\bGamma}_{XX}^{-1}{\bGamma}_{XY}. \label{eq:SolnMatrix}
\end{equation}
Moreover, the eigenvectors that correspond to the largest eigenvalues of $\bs{G}_1 \in \mathbb{R}^{P \times P}$ and $\bs{G}_2 \in \mathbb{R}^{Q \times Q}$ are defined, respectively, as
\begin{equation} \label{eq:TailTop}
    \bs{\lambda}_1 :=  \bGamma_{XX}^{1/2}\bgamma^* \in \mathbb{R}^P \ \ \text{and} \ \ \bs{\lambda}_2 := \bGamma_{YY}^{1/2}\bbeta^* \in \mathbb{R}^Q,
\end{equation}
which we refer to as the ``tail-topology''.
The absolute value of our tail-topology reflects the contribution of each component in attaining maximum tail dependence between two random vectors. In applications to brain signals, the tail-topology captures the weights assigned to each channel that drive the two brain regions to produce high amplitudes simultaneously.

\subsection{Fuzzy Clustering of Multivariate Extremes} \label{chap:subsec:fuzzyclust}

We now develop a fuzzy clustering algorithm based on CTD. One novel feature of our method is that it uses the first eigenvectors (that correspond to the largest eigenvalues), $\bs{\lambda}_1$ and $\bs{\lambda}_2$, of the canonical tail-dependence (as defined in Equation~\eqref{eq:TailTop}) to represent the topology of multivariate extremes. 
Suppose we have $N$ subjects, each with $D$-variate data $\bs{Z}_b^{(n)} \in \text{RV}^D(2)$, for $n = 1, \dots, N$, and time block $b = 1, \dots, B$.
We now provide a step-by-step procedure for Fuzzy Clustering of Multivariate Extremes via the CTD. 

\textit{Step 1: } \ \textbf{Estimate the CTD.} 
    We follow \cite{cooley2019decompositions} for estimating the TPDM. For each subject data $\{\bs{Z}^{(n)}_b\}_{b = 1}^B$, $n = 1, \dots, N$, estimate the elements of the TPDM, denoted by $\bGamma^{(n)} \in \mathbb{R}^{D\times D}$, by
    \begin{equation}
         \widehat{\Gamma}^{(n)}_{jk} = \frac{D}{c}\sum_{b=1}^B \frac{Z^{(n)}_{j,b}Z^{(n)}_{k,b}}{\{\|\bs{Z}^{(n)}_b\|\}^2}\mathbb{I}{\{\|\bs{Z}^{(n)}_b\| > r_n\}}, \label{eq:estTPDM}
    \end{equation}
    for $j,k = 1, \dots, D$, where $\mathbb{I}{\{\cdot\}}$ is the indicator function, 
    $c = \sum_{b = 1}^B \mathbb{I}{\{\|\bs{Z}^{(n)}_b\| > r_n\}}$ is the number of radial exceedances above the threshold $r_n >0$ and $\|\cdot\|$ is the L$_2$ norm. Hereafter, we take $r_n$ as the empirical $q$-th quantile, i.e., 
    \[r_n := \hat{F}^{-1}_{\|\bs{Z}^{(n)}_b\|}(q)\]
    where $\hat{F}_{\|\bs{Z}^{(n)}_b\|}(q)$ is the empirical distribution function of $\{\|\bs{Z}^{(n)}_b\|\}_{b = 1}^B$ and $q \in (0,1)$ is a probability level close to one. We advocate conducting a sensitivity analysis to choose the value of $r_n, n = 1, \dots, N$.
    Using $\widehat{\bGamma}^{(n)}$, we obtain estimates of the tail topology, $\widehat{\bs{\lambda}}_1^{(n)}$ and $\widehat{\bs{\lambda}}_2^{(n)}$, through the eigen-decomposition in Equation~\eqref{eq:SolnMatrix}.

\textit{Step 2: } \ \textbf{Stack the tail-connectivity features.} 
    Here, we take the absolute tail-topologies. \cite{lucinska2014spectral} showed that taking the absolute values of eigenvectors stabilizes clustering assignment and avoids ambiguity brought by the signs of the vectors.
    Denote the set of absolute CTD eigenvectors of $\{\bs{Z}_b^{(n)}\}_{b=1}^B$ to be 
    \[{\bs{\Lambda}_n = \left( \lvert\widehat{\bs{\lambda}}_{1}^{(n)\top}\rvert, \lvert\widehat{\bs{\lambda}}_{2}^{(n)\top} \rvert \right)^\top \in [0, \infty)^{D}}, \]
    for ${n = 1, \dots, N}$.
    Hence, for $N$ subjects, we have ${\bs{\Lambda} = \{\bs{\Lambda}_1, \dots, \bs{\Lambda}_N\} \in [0, \infty)^{D \times N}}$ as the collection of tail-topology features.

\textit{Step 3: } \ \textbf{Fuzzy clustering and computation of the membership matrix.} 
    We then perform a fuzzy $C$-means clustering (FCM) based on $\bs{\Lambda}$.
    FCM involves solving for a set of $S \in \mathbb{N}$ cluster centers 
    $\overline{\bs{\Lambda}}= \left\{\overline{\bs{\Lambda}}_{1}, \dots, \overline{\bs{\Lambda}}_{S}\right\} \in [0, \infty)^S$,
    and the fuzzy coefficient matrix, denoted as $\bs{U} \in [0,1]^{N \times S}$.
    The elements of the fuzzy coefficient matrix, $U_{ns}$, quantify the membership degree for subject $n$ to cluster $s \in \{1, \dots, S\}$. The FCM method finds $\overline{\bs{\Lambda}}$ and $\bs{U}$ by solving the minimization problem:
    \begin{equation*} \label{eq:fuzzy_opt}
        \min_{\overline{\bs{\Lambda}}, \bs{U}} \sum_{n=1}^N \sum_{s=1}^S (U_{ns})^m \left\| \bs{\Lambda}_n - \overline{\bs{\Lambda}}_{s} \right\|^2, \ \ \text{subject to } \ \sum_{s=1}^S U_{ns} = 1, \ U_{ns} \ge 0, \forall n=1,\dots,N,
    \end{equation*}
    where $m \in (1,3)$ is the fuzziness parameter that governs the extent of cluster overlap in the partition. 
    This optimization problem is solved via an iterative procedure, i.e., we update the current values of $\bs{U}$ using
    \begin{equation*}
        U_{ns} = \left( \sum_{s' \in \{1, \dots, S\}\backslash s} \left( \frac{ \left\| \bs{\Lambda}_n - \overline{\bs{\Lambda}}_{s} \right\|^2 }{ \left\| \bs{\Lambda}_n - \overline{\bs{\Lambda}}_{(s')} \right\|^2 } \right)^{\frac{1}{m-1}} \right)^{-1},
    \end{equation*}
    where $s' \neq s $. For the centers, we update them according to
    \begin{equation*}
       \overline{\bs{\Lambda}}_{s} = \frac{ W(m) \bs{\Lambda}_n}{ W(m) }, \ \ \text{where } W(m) = \sum_{n=1}^N (U_{ns})^m.
    \end{equation*}
    
\textit{Step 4: } \ \ \textbf{Label assignment.} Using the estimated membership matrix $\bs{U}$, assign a label to each subject $n = 1, \dots, N$ by selecting a cut-off. In this paper, if $U_{ns} > 0.7$, we assign subject $n$ to cluster $s$. If ${U_{ns} < 0.7}$ for all $s = 1, \dots, S$, then the subject is declared as a fuzzy-subject, i.e., the subject has tail characteristics that are compatible with more than one cluster.

For data-driven selection of $S$, we could exploit a number of cluster validity indices, e.g., the fuzzy silhouette index from \cite{rawashdeh2012crisp}. In our application, we set $S = 2$, in order to identify groups of neonates who have had seizure. We use experts' annotations as the true cluster labels and create a confusion matrix, $\bs{M} \in \{0, \dots, N\}^{2 \times 2}$, to measure accuracy. The rows of $\bs{M}$ are the experts' labels and the columns are the estimated cluster labels. We calculate the accuracy as
\begin{equation} \label{eq:accuracy}
    \text{Accuracy} = \frac{\max\{M_{11} + M_{22}, M_{12} + M_{21}\}}{N} \in [0.5,1].
\end{equation}
We perform the clustering for different values of $m$ in the interval $(1,3)$ and select the $m$ that gives the highest accuracy.


\section{Simulation} \label{chap:simulation}

In this section, we present simulation results under various settings comparing our fuzzy extremal clustering algorithm with the classical fuzzy C-means algorithm.

\subsection{Simulation Settings} \label{subchap:simsettings}
We demonstrate the accuracy of the fuzzy extremal clustering algorithm with multivariate regularly varying random vectors.
To simulate multivariate regularly varying vectors, we follow the simulation method of \cite{gong2024partial}, which is based on the framework of \cite{cooley2019decompositions}.

Let $\bs{Z}^{(n)}_b := (\bs{X}^{(n)\top}_b, \bs{Y}^{(n)\top}_b)^\top \in \RV^{12}(2)$ where $\bs{X}^{(n)}_b \in \RV^6(2)$ and $\bs{Y}^{(n)}_b \in \RV^6(2)$ represent the features from brain region 1 and brain region 2, respectively, for ${b = 1, \dots, B}$, $B = 2000$, and $n = 1, \dots, N$, with $N \in \{20,40\}$. 
Each matrix $\{\bs{Z}^{(n)}_b\}_{b = 1}^{2000}$ belongs to either cluster-1 or cluster-2. Define $d_n \in \{1,2\}$ as the cluster label for subject $n$, then $\{\bs{Z}^{(n)}_b\}_{b = 1}^{2000}$ belonging to cluster-$d_n$ is simulated via the inverse TPDM (i.e., precision matrix), $\bGamma^{-1}_{d_n} := {\bdelta_{d_n}\bdelta^\top_{d_n}}$. Hence, $\bdelta_{d_n} \in \mathbb{R}^{12 \times 12}$ is the matrix square root of the precision matrix assumed to be the same for all subjects in cluster-$d_n$.

We simulate the random vector $\bs{Z}_b^{(n)} \in \RV^{12}(2)$ using the transformed linear operations \cite{cooley2019decompositions}. Let $\ell(x) = \log(1 + \exp(x))$, $x \in \mathbb{R}$, and define the transformed linear operators as $x \circ y = \ell(x \ell^{-1}(y))$ and $y \oplus z = \ell(\ell^{-1}(y) + \ell^{-1}(z))$, $y,z \in \mathbb{R}_+$. 
Moreover, define $\bs{V}^{(n)}_b \in \mathbb{R}_+^{12}$ to be independent and identically distributed (IID) Frechet(2) random variables.
We simulate $\lfloor fN \rfloor$ fuzzy subjects, where $f \in [0,1]$.
For ``fuzzy'' subject $n$, define $F_b^{(n)} \overset{\text{IID}}{\sim} \text{Bernoulli}(0.5)$ and for ``non-fuzzy'' subject $n$, let $F_b^{(n)} = d_n - 1$ for all $b = 1, \dots, 2000$.
Then, the simulation model for subject $n$ and channel $j = 1, \dots, D$, at index $b$ is as follows:
\begin{equation}
    Z^{(n)}_{j,b} = \delta_{j1}^{(n,b)} \circ V^{(n)}_{1,b} \oplus \delta_{j2}^{(n,b)} \circ V^{(n)}_{2,b} \oplus \dots \oplus \delta_{jD}^{(n,b)} \circ V^{(n)}_{D,b}, \label{eq:simulation}
\end{equation} 
where $\delta_{jk}^{(n,b)}$ is the $(j,k)$-th entry of the matrix $\bdelta^{(n,b)} = \bdelta_{1}(1-F_b^{(n)}) + \bdelta_{2}(F_b^{(n)})$.

We then transform each $\{\bs{Z}^{(n)}_b\}_{b = 1}^{2000}$ to have unit symmetric Pareto margins with shape parameter 2, which satisfies the $\RV^1(2)$ property \citep[for details on the symmetric Pareto, see][]{jiang2025efficient}.
The analytical solution for the tail-topology of cluster $d_n$, i.e., $\bs{\Lambda}_{d_n} \in \mathbb{R}_
+^{12}$, is obtained using Proposition~\ref{prop:solnTPDM}.
The true tail-topology, $\bs{\Lambda}_{d_n}, d_n = 1,2$, is provided in Figure~\ref{Fig:SimCases}.
The accuracy of the clustering methods, given in Equation~\eqref{eq:accuracy}, is obtained for varying $N \in \{20, 40\}$, varying fuzzy parameter $m \in \{1.1, 1.2, 1.5, 1.8, 2.0, 2.2, 2.5\}$, and varying fuzzy proportion $f \in \{0, 0.1\}$.

\subsection{Simulation Results}

\begin{figure}
        \centering
        \includegraphics[width=1\linewidth]{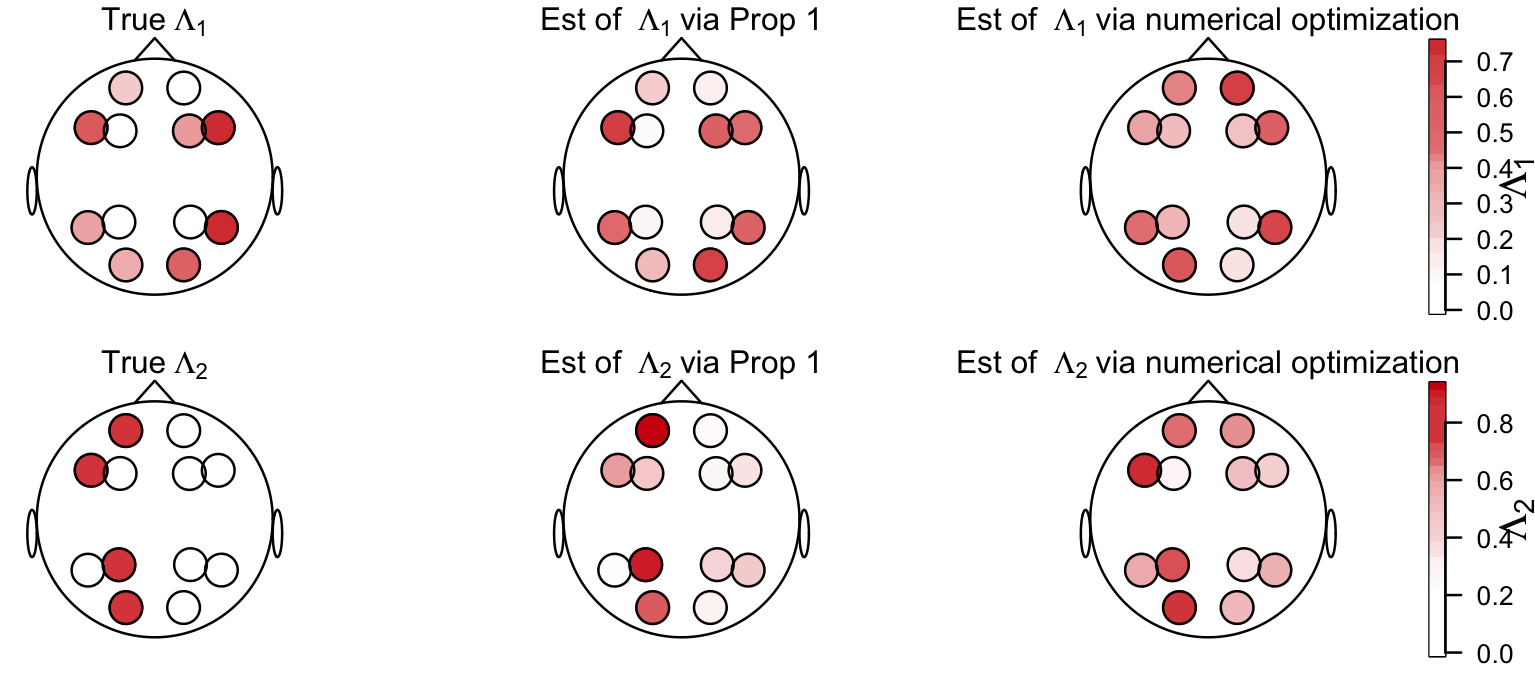}
        \caption{Comparison of $\bs{\Lambda}_1$ estimates from a single simulated $\{\bs{Z}^{(n)}_b\}_{b = 1}^{2000}$ obtained using Proposition 1 (middle panel) and direct numerical optimization (right panel) with the true value of $\bs{\Lambda}_1$ (left panel).}
        \label{Fig:SimCases}
\end{figure} 
\begin{table}
\caption{Comparison of computational cost of tail-topology estimation using Proposition 1 and standard numerical optimization.}
\label{tab:CompCost}
\begin{tabular}{@{}lrr@{}}
\toprule
Computational Cost & Estimation via Prop. 1 & Estimation via numerical optimization \\ \midrule
For $\bs{\Lambda_1}$   & 0.45 sec               & 40.83 sec                              \\
For $\bs{\Lambda_2}$   & 0.40 sec               & 41.52 sec                               \\ \bottomrule
\end{tabular}
\end{table}

\begin{figure}
        \centering
        \includegraphics[width=0.75\linewidth]{Figures/NoFuzziness-Accuracy-CTD.png}
        \caption{Accuracy of fuzzy extremal clustering (FEC) via the CTD (right) vs fuzzy C-means (FCM, left) clustering at varying levels of fuzzy parameter $m$, for $N =20$ (top) and $40$ (bottom) subjects with a proportion of fuzzy subjects equal to $f = 0$.}
        \label{Fig:Simres1}
        \includegraphics[width=0.75\linewidth]{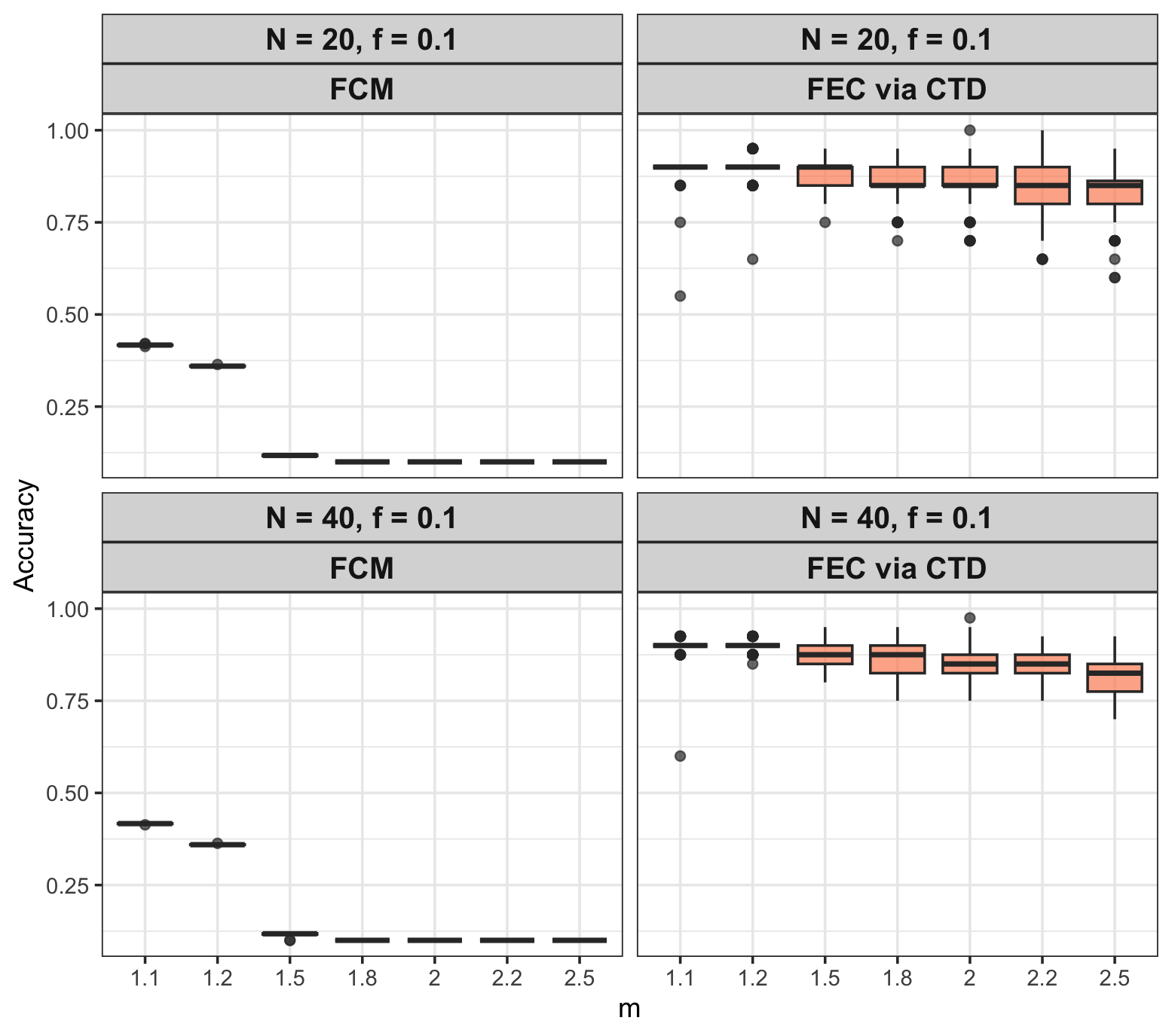}
        \caption{Accuracy of fuzzy extremal clustering (FEC) via the CTD (right) vs fuzzy C-means (FCM, left) clustering at varying levels of fuzzy parameter $m$, for $N =20$ (top) and $40$ (bottom) subjects with a proportion of fuzzy subjects equal to $f = 0.1$.}
        \label{Fig:Simres2}
\end{figure}

To empirically assess the estimated accuracy, we estimate $\bs{\Lambda}$ using two approaches: (1) the approach provided by Proposition~\ref{prop:solnTPDM}; and (2) using standard numerical optimization based on Equation~\eqref{eq:CTD}.
Figure~\ref{Fig:SimCases} compares the two approaches for estimating $\bs{\Lambda}_1$ and $\bs{\Lambda}_2$. 
From Figure~\ref{Fig:SimCases}, numerical optimization is not as accurate as the (tail-topology) estimation via the TPDM (as in Proposition~\ref{prop:solnTPDM}) when compared to the true $\bs{\Lambda}$, which is due to the fact that numerical optimization strongly depends on the choice of initial values in this high-dimensional optimization problem.
Moreover, the estimation using Proposition~\ref{prop:solnTPDM} has a much lower computational cost (see Table~\ref{tab:CompCost}). This is because the solution provided by Proposition~\ref{prop:solnTPDM} only requires estimation of the TPDM and its eigenvalue decomposition, while numerical optimization has to exhaust all possible combinations and maximizes Equation~\eqref{eq:CTD} in order to be accurate. This confirms that the approach based on Proposition~\ref{prop:solnTPDM} is both accurate and efficient.

In terms of clustering accuracy, we compared our fuzzy extremal clustering with that of a traditional FCM using the function \texttt{cmeans} in the \texttt{e1071} R-package \citep[based on work by][]{bezdek1984fcm}.
For each scenario defined in Section~\ref{subchap:simsettings}, we run the simulation study $100$ times and obtain the accuracy.
Figures~\ref{Fig:Simres1} and \ref{Fig:Simres2} summarize the $100$ results for each of the scenario. From these figures, fuzzy extremal clustering gives much higher accuracy over traditional FCM for clustering of multivariate regularly varying vectors. Moreover, increasing $N$ leads to slightly more accurate clustering. For $N = 20$ and $f = 0$, the accuracy of fuzzy extremal clustering via CTD ranges approximately from $60$\% to $100$\% for $m$ close to $1$. When $N$ is increased to $40$, the accuracy is at least $90\%$ for the same value of $m$.
Furthermore, when $f = 0$, the highest accuracy is achieved at low values of $m$, corresponding to clustering behavior that is close to hard clustering. In contrast, when $f > 0$, the maximum accuracy is reached when $m = 2$. This demonstrates the algorithm’s ability to effectively cluster data with varying levels of fuzziness. 

\begin{figure}
        \centering
        \includegraphics[width=1\linewidth]{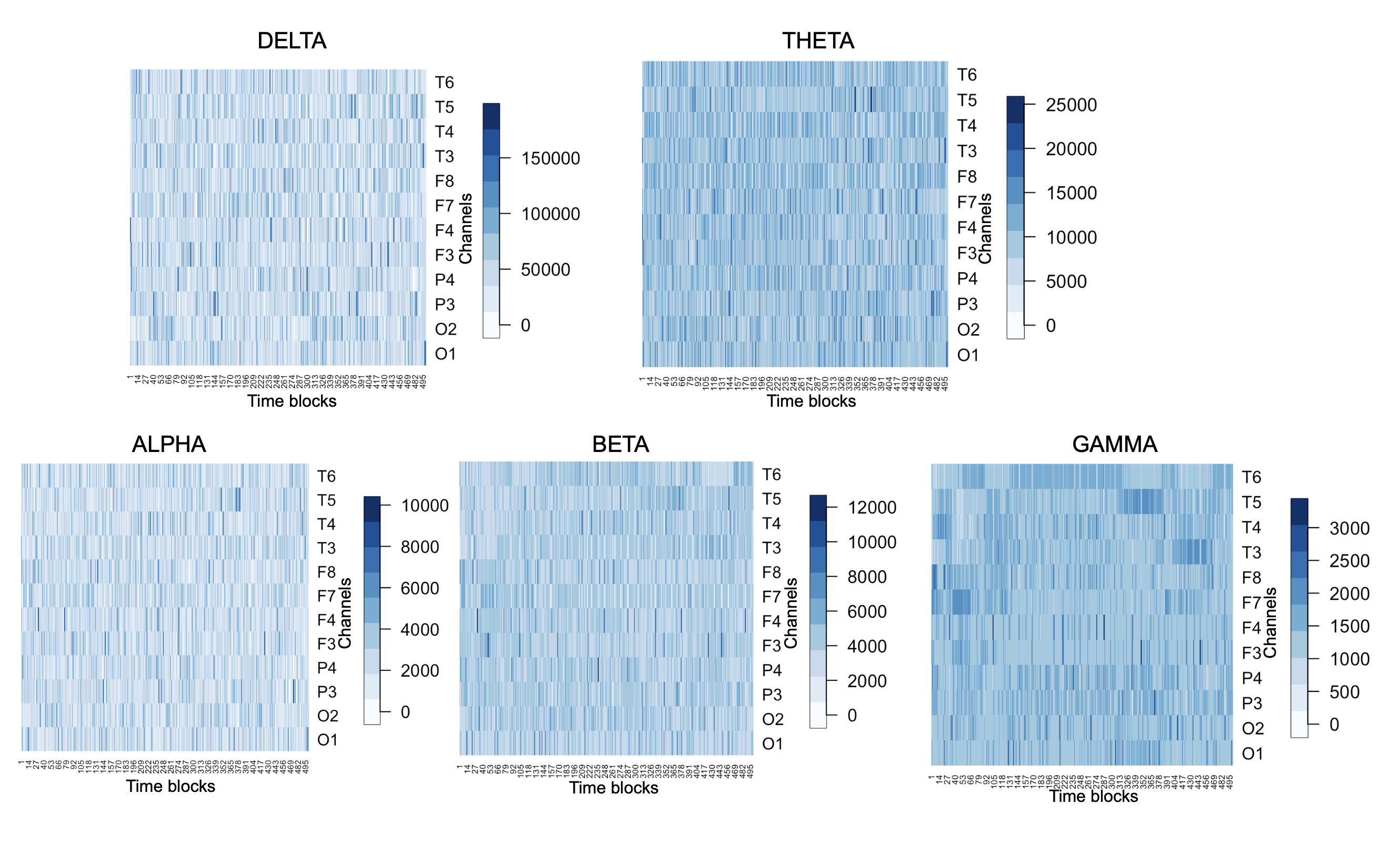}
        \caption{First 500 block-varying periodograms, $\{\bs{Z}^{(n)}_b(\Omega)\}_{b = 1}^{500}$, for a single subject $n$ and for all five frequency bands, $\Omega$, namely, delta, theta, alpha, beta, and gamma.}
        \label{Fig:Perio}
\end{figure}


\section{Extremal clustering of neonatal EEG data} \label{chap:analysis}

We apply our CTD-based extremal clustering method to neonatal EEG periodograms (Section~\ref{subsec:data}) denoted as ${\{\bs{Z}^{(n)}_b(\Omega)\}_{b = 1}^B \in \mathbb{R}_+^{D \times B}}$, for each subject $n = 1, \dots, 14$ and for each fundamental frequency band, $\Omega$, i.e., delta, theta, alpha, beta and gamma. Figure~\ref{Fig:Perio} plots the EEG periodograms of a subject for five frequency bands. From this figure, gamma-band periodograms appear to have a relatively more heavy-tailed behavior.
For each subject $n$ and a fixed $\Omega$, we standardize the data to have common margins.

\begin{figure}
        \centering
        \includegraphics[width=1\linewidth]{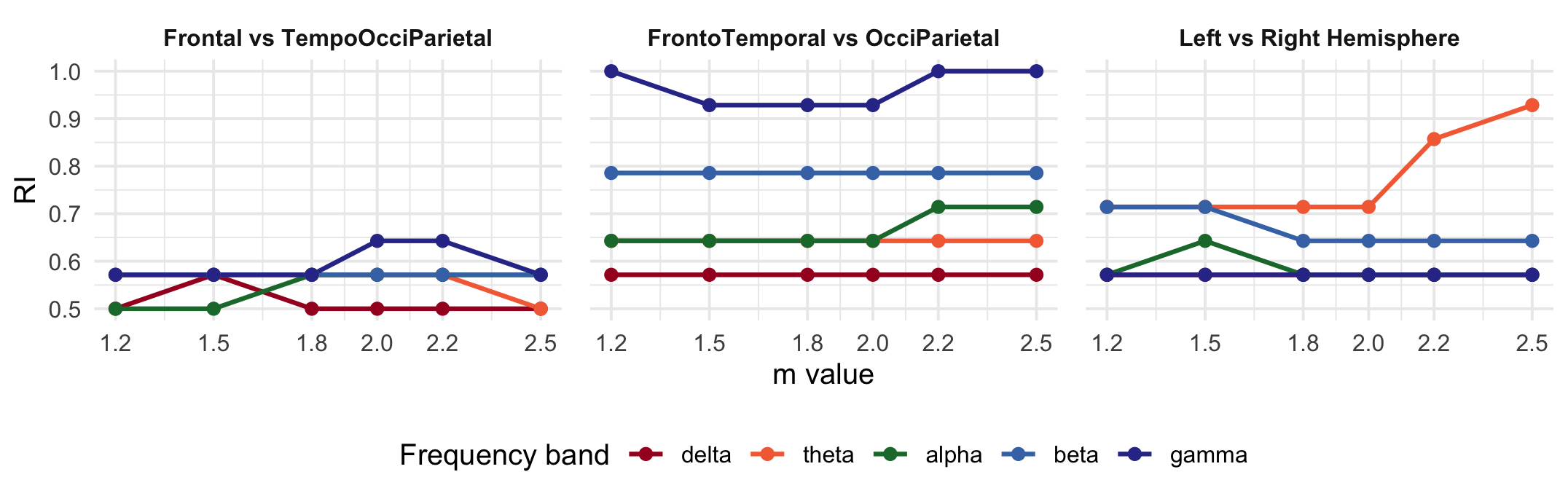}
        \caption{Accuracy of our fuzzy extremal clustering (CTD-based) evaluated for different fuzzy parameter values $m$ (x-axis), different frequency bands, $\Omega$ (colors), and different connectivity regions (panels).}
        \label{Fig:RI}
\end{figure}

\begin{figure}
        \centering
        \includegraphics[width=1\linewidth]{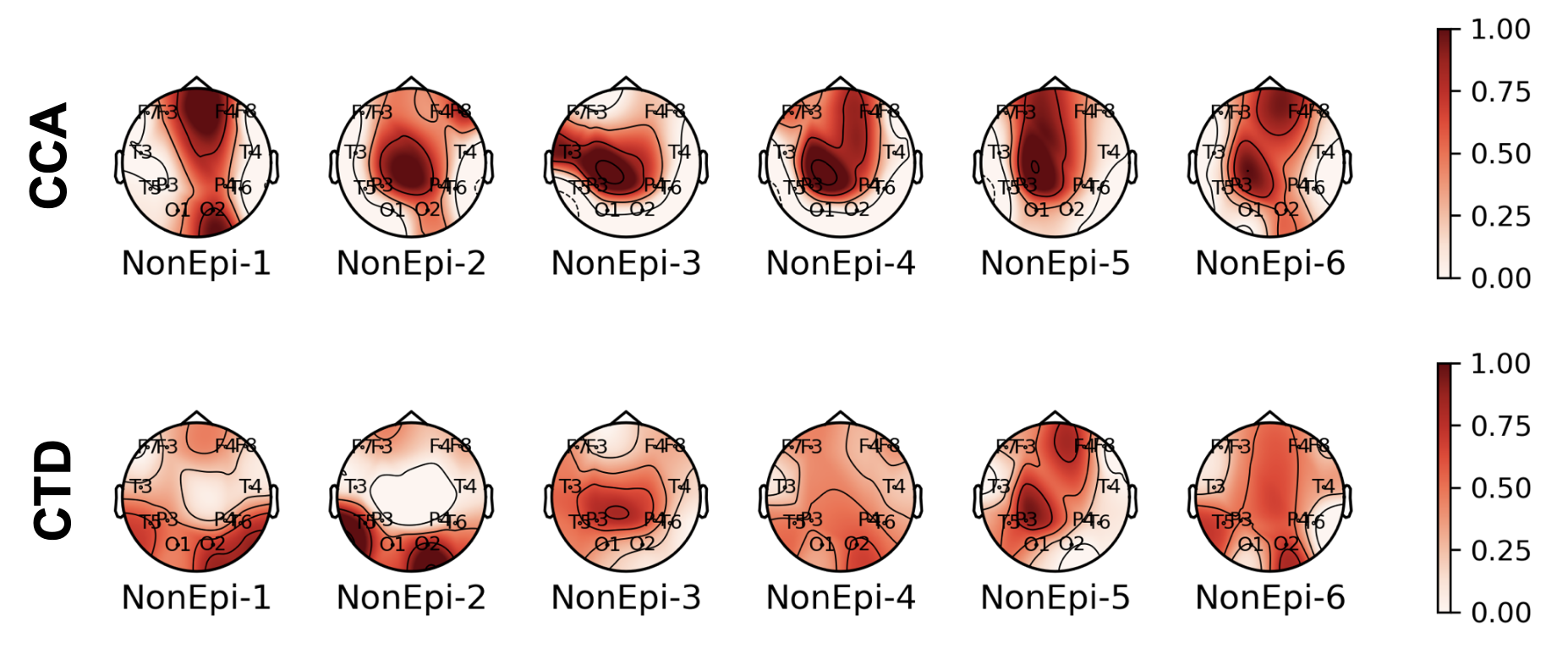}
        \caption{Tomomaps \citep[using \texttt{mne} Python package, see][]{Gursoy2014Tomopy} of CCA estimates, $|\hat{\bs{\Lambda}}_0^{(n)}|$ (first row) and tail-topology, $\hat{\bs{\Lambda}}_n$ (second row), of six non-epileptic neonates at the gamma band for FrontoTemporal-vs-OcciParietal connectivity. Darker color represents higher magnitude for channels $j = 1, \dots, 12$, labeled F3, F7, F4, F8, T3, T4, T5, T6, P3, P4, O1, O2.} 
        \label{Fig:NonEpis}
        \includegraphics[width=0.7\linewidth]{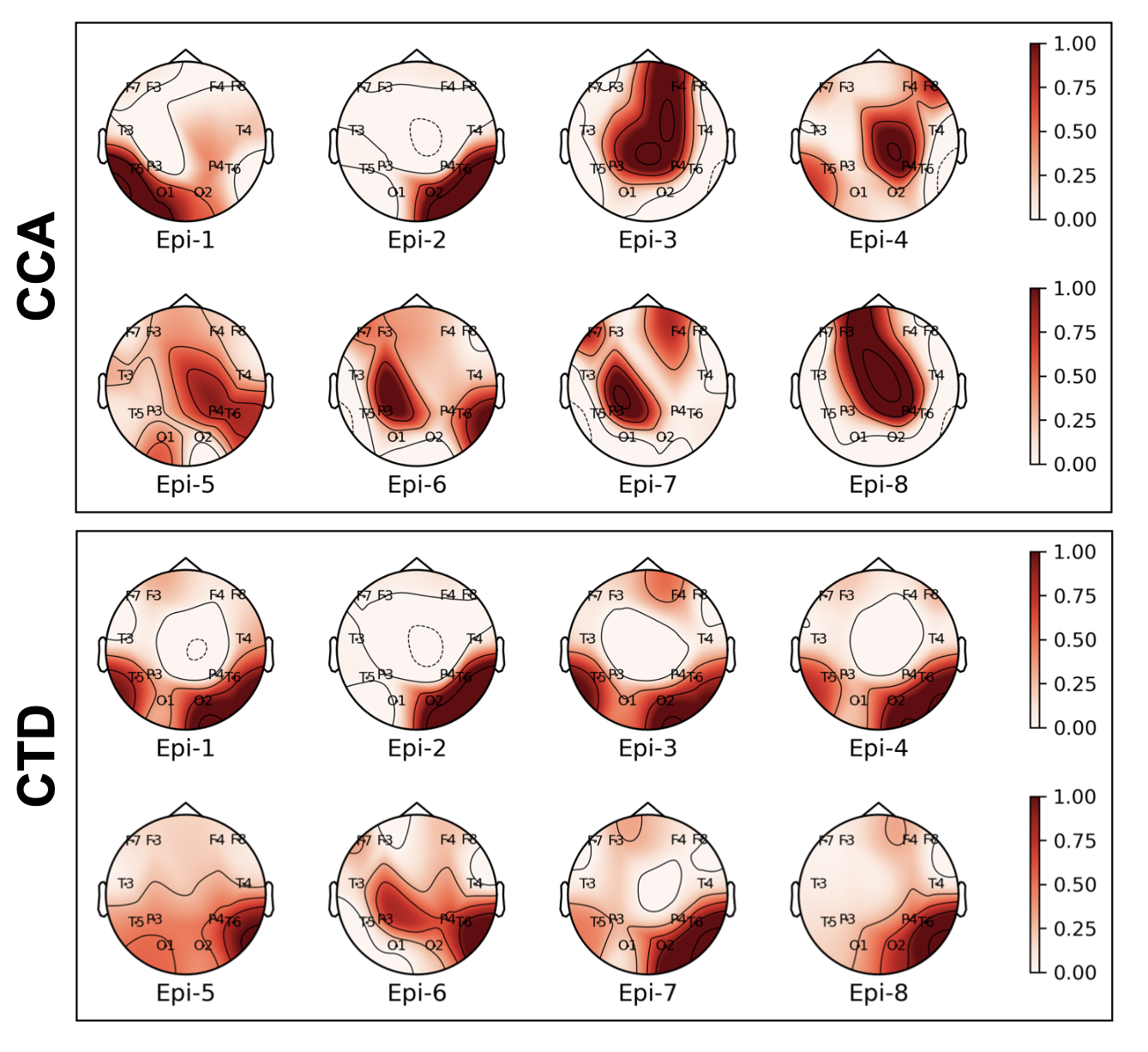}
        \caption{Tomomaps \citep[using \texttt{mne} Python package, see][]{Gursoy2014Tomopy} of CCA estimates, $|\hat{\bs{\Lambda}}_0^{(n)}|$ (first and second row), and tail-topology, $\hat{\bs{\Lambda}}_n$  (third and fourth row), of eight epileptic neonates with primary localization on \textit{right hemisphere} at the gamma band for FrontoTemporal-vs-OcciParietal connectivity. Darker color represents higher magnitude for channels $j = 1, \dots, 12$, labeled F3, F7, F4, F8, T3, T4, T5, T6, P3, P4, O1, O2.} 
        \label{Fig:Epis}
\end{figure}

\begin{figure}
        \centering
        \includegraphics[width=0.8\linewidth]{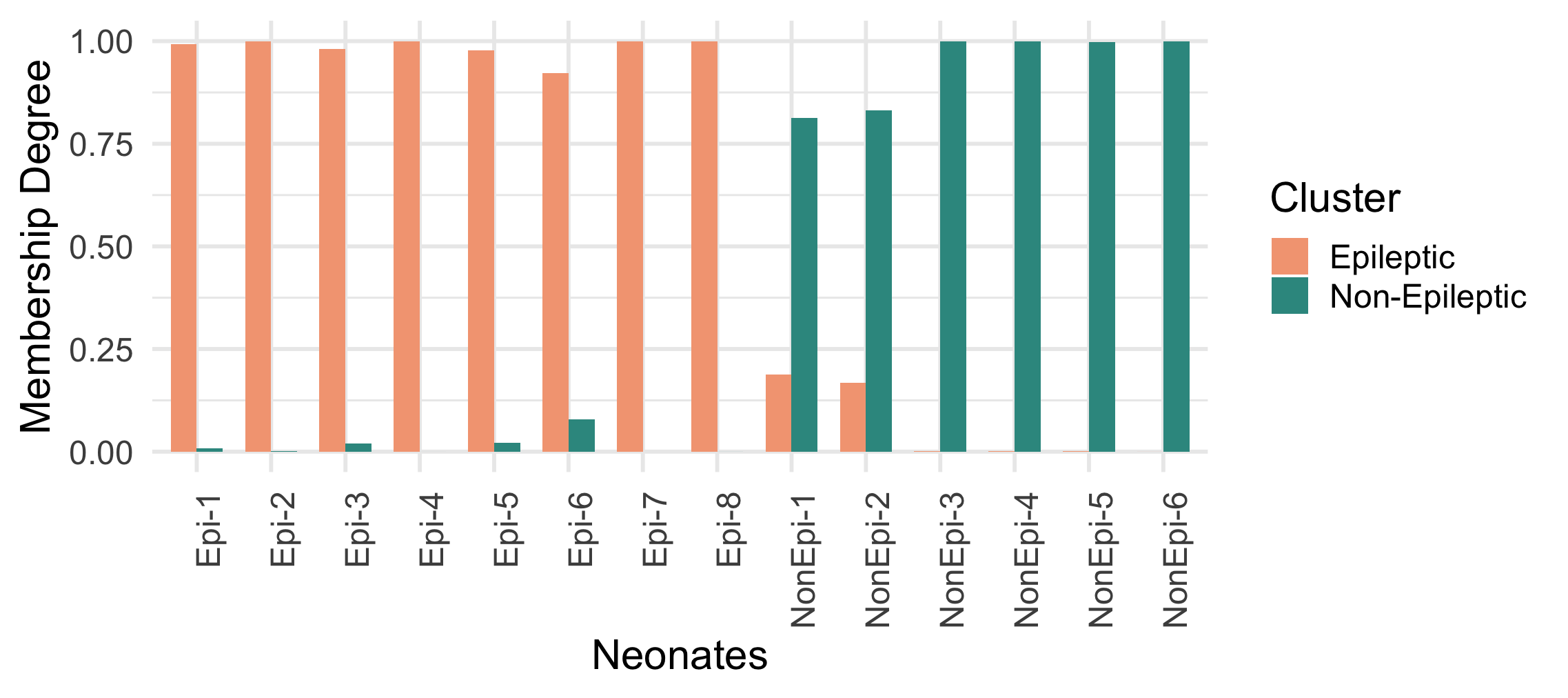}
        \caption{Membership matrix, $\bs{U}$, of our fuzzy extremal clustering for $m = 1.2$ at the gamma-band. The orange bars represent the degree of epileptic-membership $\{U_{n,1}\}_{n = 1}^{14}$, while green bars represent the degree of non-epileptic-membership of a subject, $\{U_{n,2}\}_{n = 1}^{14}$.}
        \label{Fig:MemMatrix}
\end{figure}

As EEG frequency bands are linked to human cognitive development \citep{Srinivasan2007EEG}, we aim to identify the frequency band that best distinguishes epileptic and non-epileptic neonates. Figure~\ref{Fig:RI} shows the accuracy (see Equation~\eqref{eq:accuracy}) of our clustering results based on experts annotations/labels, for each frequency band and different fuzzy-parameters $m \in \{1.2, 1.5, 1.8, 2.0, 2.5\}$.
Comparing Figures~\ref{Fig:CCA-RI} and \ref{Fig:RI} for FrontoTemporal-vs-OcciParietal and Left-vs-Right Hemisphere, our method achieves higher clustering accuracy (see Figure~\ref{Fig:RI}) than standard CCA. For FrontoTemporal-vs-OcciParietal connectivity and $m = 1.2$, our method identifies the gamma band as the best band for discriminating between epileptic and non-epileptic neonates (with an accuracy of 100\%). 
This result is in-line with the findings of \cite{guerrero2023conex} who highlighted the gamma band as the most relevant frequency for exploring tail-dependence of (seizure) EEG data. At $m = 2.5$, the CTD-based clustering of connectivity of Left-vs-Right Hemisphere at the theta-band provides more than 90\% accuracy, which implies that there are EEGs for fuzzy-neonates or neonates that exhibit tail-dependence properties of both clusters. The theta band is one of the components of burst-suppression patterns that are also found in normal EEG recordings \citep{Dulac2013, talento2025spectral}. This is one of many \textit{potential} factors that could bring out fuzziness between the clusters under Left-vs-Right Hemisphere connectivity. The Frontal-vs-TempoOcciParietal connectivity still has the lowest accuracy in discriminating epileptic and non-epileptic neonates. 

\noindent \textbf{Highlights: novel results on extremal dependence of amplitudes.} \ In Figures~\ref{Fig:NonEpis} (second row) and~\ref{Fig:Epis} (third and fourth row), we visualize the FrontoTemporal-vs-OcciParietal tail-topology at the gamma band (with $m = 1.2$) using tomomaps in \texttt{mne} Python package \citep[tomomaps in Python use Gaussian kernel-based spatial smoothing; for details of tomomaps, see][]{Gursoy2014Tomopy}. 
The tail-topology displayed in Figures~\ref{Fig:NonEpis} (second row) and~\ref{Fig:Epis} (third and fourth row) represents the absolute values of eigenvectors $\{\bs{\Lambda}_n\}_{n=1}^{14}$, with the FrontoTemporal channels populating $\{\bX_b\}_{b = 1}^{2000}$ and the OcciParietal channels populating $\{\bY_b\}_{b = 1}^{2000}$.  
Figure~\ref{Fig:Epis} (third and fourth row) shows clear and unique FrontoTemporal-vs-OcciParietal tail-topology of epileptic patients in contrast with that of non-epileptic patients (see Figure~\ref{Fig:NonEpis}, second row). The proposed method captures features about the extremes that have not been previously reported. In particular, the method demonstrates that the maximal tail-dependence between FrontoTemporal-vs-OcciParietal of epileptic patients is mostly driven by right Temporal and OcciParietal channels (i.e., P4, O6, and T6). Note that these neonates had primary localization at the right-hemisphere, and this feature is revealed only in the tails of gamma band amplitudes. Moreover, results from our proposed method confirm that the tail connectivity of neonatal EEG data from \cite{stevenson2019dataset} is distinct and more homogeneous (within clusters) compared to the bulk connectivity depicted in Figures~\ref{Fig:NonEpis} and~\ref{Fig:Epis}. Figure~\ref{Fig:MemMatrix} shows the estimated membership degree, $\bs{U} \in [0,1]^{14 \times 2}$, of each neonate for the two clusters. As $m = 1.2$, we observe that the proposed method produces a clear separation between the clusters, with only two non-epileptic neonates (NonEpi-1 and NonEpi-2) exhibiting over 10\% similarity with the epileptic cluster. The remaining subjects show near-perfect membership to the correct cluster. Referring back to Figure~\ref{Fig:NonEpis} (second row), both NonEpi-1 and NonEpi-2 are characterized by relatively substantial contributions from the Temporal and OcciParietal channels (i.e., P4, O6, and T6); however, their overall similarity to the other non-epileptic neonates remains stronger.

\section{Conclusion} \label{chap:conclusion}

In this paper, we have developed a novel method that addresses the problem of clustering extreme events---which is prevalent in many areas including
climate and finance, as well as biology and neuroscience. However, this task is non-trivial, especially when data are jointly heavy-tailed. In this paper, we extract unique features lurking at the tails of the distribution which enhance discrimination of extreme events, and paves the way for representation learning of rare events. In particular, we develop a novel way of characterizing multivariate tail-connectivity based on multivariate regular-variation \citep{resnick2007heavy}. 
Our proposed canonical tail dependence (CTD) measure is the first of its kind to offer tail-topology of connectivity between two random vectors. The CTD finds the combination of variables that provide the maximal tail-dependence between groups---we solve for unique weights attributed to each component that achieves the maximal tail-dependence of two linear combinations. These weights form the tail-topology that facilitates visualization and interpretation of clusters, highlighting a major advantage of our method.

We provide an efficient way of estimating the CTD through the tail pairwise dependence matrix \citep{cooley2019decompositions}. 
We embed the CTD estimates into the fuzzy C-means algorithm, thereby developing a novel \textit{soft clustering of multivariate extremes}. The application to neonatal EEGs revealed interesting tail-topology of epileptic and non-epileptic neonates. At the gamma band, channels in non-epileptic patients contribute equally to the maximal tail-dependence (or CTD), while few specific channels in epileptic patients are likely to be the key drivers of the CTD.  

Our method is currently limited to stationary but jointly heavy-tailed signals. Moreover, the CTD considers linear combinations of regularly varying variables. Our future work will focus on weakening the stationarity assumption by using wavelets \citep{graps1995introduction, rhif2019wavelet} as a basis, instead of Fourier coefficients. Additionally, one may also extend our approach to the case of asymptotic independence of signals, i.e., hidden regular variation \citep{resnick2002hidden}. 
Another interesting direction is to develop a data-driven approach to select the optimal groups that maximize clustering accuracy, such as fuzzy co-clustering \citep{kummamuru2003fuzzy}. Furthermore, future studies may consider the maximal extremal dependence between non-linear combinations of vectors. This could utilize deep learning which has already seen effective use in capturing complex tail-dependencies \citep[see e.g.,][]{de2025generative,shao2025modeling, hu2025gpdflow}, and has already seen use in canonical correlation analyses \citep{andrew2013deep}.


\section*{Acknowledgments}
The authors would like to thank Dan Cooley for helpful discussions and access to a working article.

\newpage
\bibliographystyle{apalike} 
\bibliography{bibliography}       

\end{document}